\documentclass[preprint,12pt]{elsarticle}

\usepackage{amssymb}
\usepackage{bm}
\usepackage{amsmath}
\usepackage{booktabs}

\journal{Pattern Recognition}

\begin{document}

\begin{frontmatter}

\title{Adaptive Conformal Inference by Particle Filtering under Hidden Markov Models}

\author[sye]{Xiaoyi Su}
\author[mn]{Zhixin Zhou}
\author[sye]{Rui Luo} %

\affiliation[sye]{organization={Department of Systems Engineering, City University of Hong Kong},%
            city={Hong Kong SAR},
            country={China}}
            
\affiliation[mn]{organization={Department of Management Sciences, City University of Hong Kong},%
            city={Hong Kong SAR},
            country={China}}

\begin{abstract}
Conformal inference is a statistical method used to construct prediction sets for point predictors, providing reliable uncertainty quantification with probability guarantees. This method utilizes historical labeled data to estimate the conformity or nonconformity between predictions and true labels. However, conducting conformal inference for hidden states under hidden Markov models (HMMs) presents a significant challenge, as the hidden state data is unavailable, resulting in the absence of a true label set to serve as a conformal calibration set.
This paper proposes an adaptive conformal inference framework that leverages a particle filtering approach to address this issue. Rather than directly focusing on the unobservable hidden state, we innovatively use weighted particles as an approximation of the actual posterior distribution of the hidden state. Our goal is to produce prediction sets that encompass these particles to achieve a specific aggregate weight sum, referred to as the aggregated coverage level. 
The proposed framework can adapt online to the time-varying distribution of data and achieve the defined marginal aggregated coverage level in both one-step and multi-step inference over the long term. We verify the effectiveness of this approach through a real-time target localization simulation study.
\end{abstract}


\end{frontmatter}

\section{Introduction}
Real-time target localization in dynamic systems presents a crucial challenge in various applications, such as autonomous driving \cite{Lu2022}, medical diagnosis \cite{VERELLEN2003129,WILLOUGHBY2006528}, and industrial automation \cite{Jondhale2022}. In these safety-critical scenarios, high accuracy alone is insufficient; reliable uncertainty quantification is also necessary to ensure the dependability of system predictions. In this context, conformal inference \cite{Gammerman1998,Vovk2005,Papadopoulos2002,Lei2013} has emerged as a promising method, providing a robust framework for integrating real-time data with probabilistic models. This method constructs prediction sets that offer probability guarantees based on statistical theory and operates without relying on distributional assumptions. The key idea is to evaluate the conformity or nonconformity between historical predictions and true labels (values) as an estimate of uncertainty in new predictions. This assessment can then be used to construct prediction sets that account for this uncertainty, with a statistically valid significance level.

Given that the true label set is a crucial input for conformal inference, the unavailability of true labels under hidden Markov models (HMMs) \cite{Leonard1966} complicates the task of predicting the target position in real time. Fortunately, particle filtering \cite{Del1996,Jun1998}, a popular and effective Bayesian approach, is employed to solve HMMs. By distributing weights among numerous particles based on corresponding likelihoods, particle filtering provides a reliable estimate of the posterior distribution.

This brings up an idea: since the actual distribution cannot be directly observed, why not capture the posterior distribution? Motivated by this idea, this paper explores the potential integration of conformal inference and particle filtering under the assumption of HMMs. The paper begins with an overview of existing conformal inference and particle filtering approaches. It then proposes an adaptive conformal inference framework based on the assumptions of hidden Markov models, utilizing the particle filtering approach. The main contributions are as follows:
\begin{itemize}
    \item \textit{Utilization of Weighted Particles in Conformal Inference:} Under the hidden Markov model, the proposed method novelly employs the numerous weighted particles generated by particle filtering as an estimate of the posterior distribution, incorporating them into conformal inference. Here, coverage is defined as the sum of the weights of the particles contained in the predicted set, thereby addressing the challenges posed by hidden states.
    \item \textit{Adaptive Online Update Strategy:} The proposed conformal inference framework features an online update strategy, making it adaptive to the time-shifting distributions of data while maintaining a reliable convergence property of coverage.
    \item \textit{Multi-Step Conformal Inference:} The proposed method can be extended to multi-step inference, robustly achieving the marginal coverage level at each step.
\end{itemize}
The organization of this paper is as follows: Section \ref{sec: relatedwork} provides a brief summary of related research work. Section \ref{sec: methodology} introduces the components of the proposed adaptive conformal inference framework. The performance of this proposed framework is then evaluated through a real-time target localization simulation study in Section \ref{sec: simulationstudy}, and the findings are discussed in Section \ref{sec: discussion}. Finally, Section \ref{sec: conclusion} concludes the paper.

\section{Related Work}
\label{sec: relatedwork}
This section summarizes existing conformal prediction methods and particle filtering approaches, highlighting their characteristics, followed by a driving example to illustrate the proposed method in this paper.

Conformal prediction (CP) has been widely applied as a method for obtaining reliable estimates of prediction confidence. Fontana et~al.~\cite{Matteo2023} provided a comprehensive review of conformal inference and its novel developments, covering the theoretical basis and more advanced adaptations. In recent years, with advancements in computational techniques, conformal inference has gained popularity in classification \cite{luo2024trustworthy, luo2024weighted} and regression \cite{luo2024conformal, luo2024conformalized} problems, integrating with statistical learning methods and machine learning algorithms \cite{Anastasios2023}. In particular, Stutz et~al.~\cite{stutz2023conformal} proposed the Monte Carlo conformal prediction method for classification tasks with ambiguous ground truth labels, using the aggregated distribution of voted labels by experts as an approximation to the distribution of ground truth labels.

Gibbs and Cand{{\`e}}s~\cite{Gibbs2021} developed adaptive conformal inference (ACI) methods in an online setting, providing distribution-free theoretical results. Their methods involve rolling estimates of the quantile function of conformity scores and an online update strategy for the argument to the quantile function. However, the choice of step size parameter $\gamma$ in this update strategy entails a trade-off between adaptability and stability. Zaffran et~al.~\cite{Chaudhuri2022} introduced online expert aggregation on ACI (AgACI), aiming to learn this $\gamma$ within ACI. Moreover, Bastani et~al.~\cite{bastani2022practical} instead considered selecting the most promising argument for the quantile function within a grid to achieve the desired coverage level over the long term using the multivalid predictor (MVP) algorithm. By modifying the algorithm from Gradu et~al.~\cite{Gradu2023}, Gibbs and Cand{{\`e}}s~\cite{Gibbs2024} updated their ACI to the dynamically-tuned adaptive conformal inference (DtACI), which runs multiple versions of ACI in parallel with a set of candidate step size parameters, choosing the argument based on historical performance. Additionally, Cleaveland et~al.~\cite{Cleaveland2024} recently proposed an optimization-based method to reduce the conservatism in conformal inference for time series data. Nettasinghe et~al.~\cite{Nettasinghe2023}
extended the conformal inference to hidden Markov models but assumed available calibration set of hidden state.

Particle filtering (PF) is a sequential Monte Carlo method \cite{Jun1998} that utilizes plenty of weighted particles in variable space to estimate the actual posterior distribution, with weights proportional to the likelihood given by observational data. Gustafsson et~al.~\cite{Gustafsson2002} provided a comprehensive overview of sequential Monte Carlo methods, discussing their theoretical foundations and significant practical advancements. By including a resampling step, i.e., drawing samples from the weighted particles according to the normalized weights, particle filtering can avoid the potential weight collapse due to disparities in weights, which is known as the sequential importance resampling method \cite{Doucet2000}. Pitt and Neil \cite{Pitt1999} proposed the auxiliary particle filtering method, which samples from the joint conditional density of the state and the indices of particles, where each index refers to an auxiliary variable that is discarded after sampling. This high-dimensional sampling method yields better samples and reduces the required number of particles, especially in the presence of outliers in the conditional distribution. Djuri{\'{c}} et~al.~\cite{Djuric2008} utilized the auxiliary particle filtering approach to design an algorithm for single target tracking in a binary sensor network. Additionally, Ren et al.~\cite{ren2023conditional} introduced a fixed-time conditional integral sliding mode control strategy for stabilizing fuzzy uncertain complex systems, demonstrating robust uncertainty management and rapid stabilization. 

Similar to \cite{Djuric2008},  we consider a binary sensor network model to detect a moving target in the environment as an example scenario to explain our method. 
Under the assumption of hidden Markov models (HMMs), conducting conformal inference for the hidden state is challenging, as empirical coverage cannot be directly obtained. However, the observational data from the sensor network allows the particle filtering method to approximate the posterior distribution of the target position. Using these weighted particles, we conduct conformal inference to construct prediction sets that encompass enough particles to achieve a specific level of weighted sum. In this way, we can obtain prediction regions for the target position, even in the absence of actual position data.

\section{Methodology}
\label{sec: methodology}
In this section, we introduce the proposed framework that integrates hidden Markov models (HMMs), particle filtering, and adaptive conformal inference. First, we establish several assumptions regarding the environmental model utilized to further elucidate our framework. Next, we review general particle filtering and auxiliary particle filtering approaches, followed by explaining the specific adaptive conformal inference method, which serve as the foundations of our framework. Finally, we demonstrate how the proposed framework can be extended to conduct multi-step inference.

\subsection{Target Motion Model}
\label{subsec: motionmodel}
We assume that the target motion follows a standard model commonly utilized in the literature, as noted in \cite{Djuric2008, Gustafsson2002}, which has the form
\begin{equation}\label{eq: motionmodel}
\bm{X}_{t+1} = \bm{P}\bm{X}_{t} + \bm{Q}\bm{a}_{t},
\end{equation}
where $\bm{X}_t = \left(x_{1,t}, x_{2,t}, v_{1,t}, v_{2,t}\right)'$, $\bm{a}_{t} = \left(a_{1,t}, a_{2,t}\right)' \sim N\left(\bm{0}, \Sigma\right),\ \Sigma = \text{diag}(\sigma_1^2, \sigma_2^2)$, 
\begin{equation}
\bm{P} = 
\begin{pmatrix}
1 & 0 & \text{dt} & 0\\
0 & 1 & 0 & \text{dt}\\
0 & 0 & 1 & 0\\
0 & 0 & 0 & 1
\end{pmatrix},\ 
\bm{Q} = 
\begin{pmatrix}
(\text{dt})^2/2 & 0 \\
0 & (\text{dt})^2/2 \\
\text{dt} & 0 \\
0 & \text{dt} 
\end{pmatrix}.
\end{equation}
$\bm{X}_t$ denotes the state of the target at time~$t$, with the first two components $(x_{1,t}, x_{2,t})$ representing the spatial coordinates and the last two components $(v_{1,t}, v_{2,t})$ indicating the corresponding velocities. The variable $\bm{a}_{t}$ is a 2-dimensional Gaussian random variable representing the acceleration of the target in the respective dimensions. The matrices $\bm{P}$ and $\bm{Q}$ reflect the general laws of physics governing the target's motion. The variable $\text{dt}$ represents the pre-set sampling period.

This model captures the inherent uncertainties in target motion, accounting for realistic scenarios where external forces and environmental factors may lead to random acceleration, thus affecting velocity. The choice of this model is justified by its capability to represent realistic motion patterns and its compatibility with the particle filtering approach utilized for our adaptive conformal inference.

\subsection{Sensor Network Model}
To detect the target, we assume a sensor network model that captures the target's information within a specified area, yielding binary outcomes. We explain how the sensors are deployed in the map and illustrate the detection mechanism linking the sensors to the target.

The deployment of sensors is crucial for optimizing coverage and detection performance. For convenience, we assume a randomized placement strategy to distribute sensors across the operational area surrounding the target's actual positions. Specifically, let $d_s$ denote the density of sensors and $A_{map}$ represent the area of the map under study. The number of sensors $n_s$ is:
\begin{equation}
n_s = d_s \cdot A_{map}.
\end{equation}
The spatial coordinates of all sensors are assumed to be independent and identically distributed (i.i.d.) uniform samples from the map.

At each time~$t$, sensor~$i$ provides a binary outcome $y_{i,t} \in \{0,1\}$, indicating the successful detection of the target $(y_{i,t}=1)$ or failure $(y_{i,t}=0)$. It is reasonable to assume the detection probability is influenced by the distance between each sensor and the target, with closer sensors having a higher probability of detection. 
Within a specific range, signal transmission is assumed to be strong, facilitating target detection. Thus, the sensors' detection capabilities are expressed by the following probabilistic model, which accounts for both target proximity and potential strong signal range effects:
\begin{equation}
\mathbb{P}\left(y_{i,t}=1\right) 
= w \cdot \exp\left({-\beta\cdot d_i^2}\right) 
+ (1-w)\cdot p_0 \cdot \mathbb{I}(d_i\le r_0),
\end{equation}
where $d_i$ is the distance between sensor~$i$ and the target, $\beta > 0$ is a decay parameter affecting the influence of distance, $r_0$ represents the strong signal range and $p_0 \in [0,1]$ is the basic detection probability within this range. We also introduce a weight parameter $w \in [0,1]$ to balance these two types of influences. 

Under this assumed sensor network model structure, we can obtain an observed vector $\bm{Y}_t = \left(y_{1,t}, y_{2,t}, \dots, y_{n_s,t}\right)'$ at each time~$t$. By combining the target motion model and the sensor network model, we construct a hidden Markov model for the target motion:
\begin{equation}
\left(\bm{X}_t, \bm{Y}_t\right), \ t=1,2,\dots,
\end{equation}
in which $\bm{X}_t$ is the hidden state vector and $\bm{Y}_t$  is the observed vector.

\subsection{Particle Filter Predictor}\label{sec: predictor}
Particle Filtering is a widely used Bayesian approach for addressing filtering problems in dynamic systems. Specifically, 
it is a powerful sequential Monte Carlo method that estimates the state of a dynamic system by representing the posterior distribution with a set of particles. This method exhibits significant advantages in scenarios where the system dynamics and observation models are non-linear and non-Gaussian, making it particularly suitable for target tracking in dynamic environments. In addition to the general sequential importance resampling (SIR)-based particle filtering (PF), we also introduce the auxiliary particle filtering (APF) method proposed by \cite{Pitt1999}, which has been thoroughly explained \cite{Doucet2001SequentialMC} and applied to address target tracking problems in binary sensor networks \cite{Djuric2008}.

\subsubsection{PF Predictor}
We first consider the general procedures of particle filtering. At each time~$t$, given $\bm{Y}_{1:t} := \{\bm{Y}_1,\bm{Y}_2,\dots, \bm{Y}_t\}$, which represents the historical observations up to time~$t$, the posterior distribution of the hidden state$\bm{X}_t$, denoted as $p\left(\bm{X}_t|\bm{Y}_{1:t}\right)$, is approximated by a set of particles. 
Let $M$ be the number of particles, and let $\{(\bm{X}_t^{(j)}, w_t^{(j)})\}_{j=1}^{M}$ denote the particle set at time~$t$, where $\bm{X}_t^{(j)}$ and $w_t^{(j)}$ indicate the state and weight of particle~$j$, respectively. Then the estimated distribution
\begin{equation}
\hat{p}\left(\bm{X}_t|\bm{Y}_{1:t}\right)
= \sum_{j=1}^{M}\Tilde{w}_t^{(j)}\delta_{\bm{X}_t^{(j)}}(\bm{X}_t), \ 
\Tilde{w}_t^{(j)}= \frac{w_t^{(j)}}{\sum_{k=1}^{M}w_t^{(k)}},
\end{equation}
serves as an approximation to $p\left(\bm{X}_t|\bm{Y}_{1:t}\right)$ \cite{Hu2008}. Here, $\delta_{\bm{X}_t^{(j)}}(\cdot)$ is the Dirac delta function, which assigns mass at the point $\bm{X}_t^{(j)}$, $j=1,\dots,M$.

The general procedures of particle filtering can be described as follows:
\begin{itemize}
    \item \textbf{Initialization:} Given a prior distribution $\pi_0$, initialize the particle set by sampling $\bm{X}_0^{(j)} \sim \pi_0$ and setting $w_0^{(j)}=\Tilde{w}_0^{(j)}=1/M$, $j=1,2,\dots,M$.
    \item \textbf{New Particles Generation:} Generate new particles based on the motion model (\ref{eq: motionmodel}), i.e., 
    \begin{equation}
        \bm{X}_{t}^{(j)} = \bm{P}\bm{X}_{t-1}^{(j)} + \bm{Q}\bm{a}_{t-1}^{(j)},\ j=1,\dots,M.
    \end{equation}
    \item \textbf{Weights Update:} Update and normalize the weights according to the observation model by calculating the likelihood:
    \begin{equation}
        w_{t}^{(j)} = p\left(\bm{Y}_{t}|\bm{X}_{t}^{(j)}\right),\ \Tilde{w}_t^{(j)}= \frac{w_t^{(j)}}{\sum_{k=1}^{M}w_t^{(k)}},
    \end{equation}
    where $j=1,\dots,M$.
    \item \textbf{Resampling:} Draw $M$ samples from the current particle set with replacement, in accordance with their corresponding normalized weights $\Tilde{w}_t^{(j)}$, $j=1,2,\dots,M$.
    \item \textbf{Iteration:} Iteratively repeat the above steps (excluding initialization) for each subsequent time step $t$, until the final time step $T$ is reached.
\end{itemize}
The key idea of our PF predictor is to incorporate a point prediction step after the generation of new particles. Specifically, before obtaining the observation vector $\bm{Y}_{t}$, we provide a one-step-ahead prediction of the hidden state $\bm{X}_{t}$:
\begin{equation}
\hat{\bm{X}}_{t} := f_{PF}\left(\{\bm{X}_{t}^{(j)}\}_{j=1}^{M}\right) 
= \frac{1}{M}\sum_{j=1}^{M}\bm{X}_{t}^{(j)},
\end{equation}
which can also be expressed as
\begin{equation}
\hat{\bm{X}}_{t} = f_{PF}\left(\bm{Y}_{1:t-1}\right).
\end{equation}

\subsubsection{APF Predictor}
The auxiliary particle filtering is a modified version of particle filtering that incorporates auxiliary variables to perform filtering in a higher dimension. The procedures are outlined as follows:
\begin{itemize}
    \item \textbf{Initialization:} Given a prior distribution $\pi_0$, initialize the particle set by sampling $\bm{X}_0^{(j)} \sim \pi_0$ and setting $w_0^{(j)}=1/M$, $j=1,2,\dots,M$.
    \item \textbf{Particle Streams Selection:} Sample $M$ indices of particles from the particle set with replacement according to the probabilities:
    \begin{equation}
        \Bar{w}_{t}^{(j)} \varpropto p\left(\bm{Y}_{t}|\bm{\mu}_{t}^{(j)}\right)\cdot w_{t-1}^{(j)},
    \end{equation}
    where $\sum_{j=1}^{M}\Bar{w}_{t}^{(j)}=1$. $\bm{\mu}_{t}^{(j)}$ is the conditional mean of $\bm{X}_{t}^{(j)}$, given by
    \begin{equation}
        \bm{\mu}_{t}^{(j)} = E\left(\bm{X}_{t}^{(j)}|\bm{X}_{t-1}^{(j)}\right) = \bm{P}\bm{X}_{t-1}^{(j)}.
    \end{equation}
    Let $\{k_j\}_{j=1}^{M}$ denote the resulting indices.

    \item \textbf{New Particles Generation:} Generate new particles using the following equation:
    \begin{equation}
        \bm{X}_{t}^{(j)} = \bm{P}\bm{X}_{t-1}^{(k_j)} + \bm{Q}\bm{a}_{t-1}^{(k_j)},\ j=1,\dots,M.
    \end{equation}
    \item \textbf{Weights Update:} update and normalize the weights by 
    \begin{equation}
        w_{t}^{(j)} \varpropto \frac{p\left(\bm{Y}_{t}|\bm{X}_{t}^{(j)}\right)}{p\left(\bm{Y}_{t}|\bm{\mu}_{t}^{(k_j)}\right)},\ j=1,\dots,M.
    \end{equation}
    where $\sum_{j=1}^{M}w_{t}^{(j)}=1$.
    \item \textbf{Iteration:} Iteratively repeat the above steps (excluding initialization) until the final time step $T$ is reached.
\end{itemize}
Similarly, we construct our APF predictor by adding a point prediction step after the generation of new particles to obtain
\begin{equation}
\hat{\bm{X}}_{t} := 
f_{APF}\left(\{\bm{X}_{t}^{(j)}\}_{j=1}^{M}\right)
= \frac{1}{M}\sum_{j=1}^{M}\bm{X}_{t}^{(j)}.
\end{equation}

\subsection{Adaptive Conformal Inference}
Let $\bm{x}_{t}=(x_{1,t}, x_{2,t})'$ and $\bm{v}_{t}=(v_{1,t}, v_{2,t})'$ denote the spatial coordinate and velocity components of the target, respectively. At time step $t$, the previously mentioned PF and APF predictors provide an online one-step-ahead point prediction for the hidden state $\bm{X}_{t+1} = (\bm{x}_{t+1}', \bm{v}_{t+1}')'$, particularly for the position of the target, $\bm{x}_{t+1}$. By leveraging these two predictors, we aim to construct conformal inference for the target position.

A typical conformal inference framework includes a dataset split into a training set and a calibration set, a predictive model, a conformity score function, and a target coverage level for the predicted set to encompass the actual variable of interest. When dealing with time series data, the training and calibration sets usually originate from the historical data $\{\bm{X}_{1:t}, \bm{Y}_{1:t}\}$. However, under the assumption of a hidden Markov model, the hidden state vector $\bm{X}$ cannot be observed directly; all related information is derived from the observation vector $\bm{Y}$. Assuming the coverage level is set to $1-\alpha$, we consider the objective of constructing a predicted set for $\bm{x}_{t+1}$ that is as small as possible while still covering $1-\alpha$ of the density of the posterior distribution $p(\bm{x}_{t+1}|\bm{Y}_{1:t+1})$ at time step $t$. Since $p(\bm{x}_{t+1}|\bm{Y}_{1:t+1})$ is a continuous distribution over the 2-dimensional space, its density is challenging to compute directly. Instead, we treat the particle set $\{(\bm{x}_{t+1}^{(j)}, w_{t+1}^{(j)})\}_{j=1}^{M}$ as the weighted samples from $p(\bm{x}_{t+1}|\bm{Y}_{1:t+1})$. Thus, the posterior distribution is approximated by a discrete distribution with a probability mass function ($pmf$) given by
\begin{equation}\label{eq: aggregateddistribution}
\mathbb{P}(\bm{x}_{t+1} = \bm{x}_{t+1}^{(j)}|\bm{Y}_{1:t+1})=w_{t+1}^{(j)},\ j=1,2,\dots,M, 
\end{equation}
where $\sum_{j=1}^{M}w_{t+1}^{(j)}=1$. Consequently, the predicted set must cover the particles such that the sum of their weights is not less than $1-\alpha$. Using the terminology from \cite{stutz2023conformal}, we refer to this sum of weights as ``aggregated coverage", and the distribution (\ref{eq: aggregateddistribution}) defined above as the ``aggregated posterior distribution".

We choose either the PF predictor or the APF predictor as the predictive model, denoted as $f(\cdot)$, which provides
\begin{equation}
    f\left(\{\bm{x}_{t+1}^{(j)}\}_{j=1}^{M}\right) = \hat{\bm{x}}_{t+1}.
\end{equation}
To quantify the conformity of any selected spatial coordinate $\bm{x}$ with the predicted position $\hat{\bm{x}}_{t+1}$ given by the predictor $f$, we define a conformity score function $C(\cdot, \cdot)$ as
\begin{equation}
    C(\hat{\bm{x}}_{t+1}, \bm{x}) := \|\hat{\bm{x}}_{t+1} - \bm{x}\|, 
\end{equation}
where $\|\cdot\|$ denotes the $L^2$ norm, measuring the Euclidean distance between two points. The primary concern is how to determine an appropriate threshold for $C(\hat{\bm{x}}_{t+1}, \bm{x})$ to include $\bm{x}$ in the predicted set, ensuring that the set size is minimized while maintaining aggregated coverage rate at the level $1-\alpha$. 

Let $\mathcal{C}_{t}(b)\ (1 \le b <t)$ denote a calibration set at time $t$ that contains the predicted position data and weighted particle data from the most recent $b$ time steps. Specifically, we define:
\begin{equation}\label{eq: calibrationset1}
   \mathcal{C}_{t}(b) := 
   \left\{
   \hat{\bm{x}}_{t-b+1:t},\ (\bm{x}_{t-b+1:t}^{(j)}, w_{t-b+1:t}^{(j)})
   \right\}_{j=1}^{M}.
\end{equation}
This calibration set $\mathcal{C}_{t}(b)$ forms a discrete probability distribution for the conformity scores between the predicted position and the weighted particles. For any $t-b+1\le k \le t$, the conformity score $C(\hat{\bm{x}}_{k}, \bm{x}_{k}^{(j)})$ has a probability mass of $w_{k}^{(j)}/(\sum_{l,i}w_{l}^{(i)}) = w_{k}^{(j)}/b$, where $t-b+1\le l \le t$ and $1\le i \le M$. We can define the estimated quantiles of the conformity scores at time $t+1$ as
\begin{equation}
    \hat{\mathcal{Q}}_{t+1}(q) 
    := \inf\left\{
    c: \sum_{k,j}
    \mathbb{I}\left(\delta_{k}^{(j)} \le c \right)
    \cdot \frac{w_{k}^{(j)}}{b}
    \ge q
    \right\},
\end{equation}
where $\delta_{k}^{(j)} = C(\hat{\bm{x}}_{k}, \bm{x}_{k}^{(j)})$, $t-b+1\le k \le t$, $1\le j \le M$. A natural definition of the predicted set is then given by
\begin{equation}\label{eq: naturalpredictedset}
    \hat{\mathcal{X}}_{t+1}(\alpha) :=
    \left\{\bm{x}: 
    C(\hat{\bm{x}}_{t+1}, \bm{x}) \le \hat{\mathcal{Q}}_{t+1}(1-\alpha)
    \right\}.
\end{equation}

However, since the distribution of data generally shifts over time, the estimated $\hat{\mathcal{Q}}_{t+1}$ is unstable. As a result, the realized aggregated miscoverage rate of the predicted set (\ref{eq: naturalpredictedset}) at time $t+1$, defined as 
\begin{equation}
    \Lambda_{t+1}(\alpha) 
    = \sum_{j=1}^{M} w_{t+1}^{(j)} \cdot 
    \mathbb{I}\left(
    \delta_{t+1}^{(j)}>\hat{\mathcal{Q}}_{t+1}(1-\alpha)
    \right),
\end{equation}
may deviate significantly from the level $\alpha$. Therefore, to achieve the aggregated miscoverage level $\alpha$, the predicted set should be adjusted based on the parameter $\alpha$ in (\ref{eq: naturalpredictedset}) to adapt to the shifting data distribution. We consider a similar adaptive strategy for conformal inference as proposed by \cite{Gibbs2021}. 

Let $C^{*}>0$ be a sufficiently large constant such that for any $1\le j \le M$, $\mathbb{P}(\delta_{t+1}^{(j)}>C^{*})=0$. We further assume that $\hat{\mathcal{Q}}_{t+1}(q) = 0$ for all $q\le 0$ and $C^{*}$ for all $q\ge 1$. Then, there exists
\begin{equation}
    \alpha_{t+1}'=\sup\arg\limits_{\beta\in[0,1]}
    \left\{
    \Lambda_{t+1}(\beta)\le \alpha
    \right\},
\end{equation}
which makes the realized aggregated miscoverage rate as close as possible to the level $\alpha$ based on the estimated $\hat{\mathcal{Q}}_{t+1}$. Therefore, it is possible to reach the aggregated miscoverage level $\alpha$ by adjusting the input parameter to $\hat{\mathcal{Q}}_{t+1}$. Specifically, we first initialize the $\alpha_1$ to $\hat{\mathcal{Q}}_{1}$. For instance, we can set $\alpha_1 = \alpha$. Then given a step size $\gamma>0$, the parameter is updated online by
\begin{equation}
    \alpha_{t+1} = \alpha_{t} + \gamma\left(\alpha - \Lambda_{t}(\alpha_{t})\right).
\end{equation}
Although the use of $\Lambda_{t}(\alpha_{t})$ to update $\alpha_{t+1}$ differs from the approach in \cite{Gibbs2021}, it is not difficult to derive some similar properties. For example, we have the property:
\begin{equation}\label{eq: property1}
    \mathbb{P}\left(\alpha_{t} \in [-\gamma, 1+\gamma],\ \forall\ t\in \mathbb{N}\right) = 1,
\end{equation}
and a convergence property
\begin{equation}\label{eq: property2}
    \mathbb{P}\left(
    \lim\limits_{T\to \infty}
    T^{-1}\sum_{t=1}^{T}\Lambda_{t}(\alpha_{t})
    =\alpha
    \right) = 1.
\end{equation}
This indicates that $\Lambda_{t}(\alpha_{t})$, the mean aggregated miscoverage rate of the predicted set $\hat{\mathcal{X}}_{t}(\alpha_{t})$, will converge to the level $\alpha$ in the long term.

\subsection{Multi-Step Inference Strategy}
In Section \ref{sec: predictor}, we introduce two predictors: the PF predictor and the APF predictor, both of which provide one-step-ahead predictions for the hidden state $\bm{X}$ by propagating the particles according to the motion model (\ref{eq: motionmodel}). One idea is that if the particle set accurately captures the motion pattern of the target at time $t$, the motion of the particle set will not deviate significantly from the actual target motion over a short period in the future. Thus, it is reasonable to propagate the particles for multiple time steps to obtain multi-step predictions from the predictors, followed by extending the online adaptive conformal inference to multi-step inference.

Specifically, let $\{(\bm{X}_{t}^{(j)}, w_{t}^{(j)})\}_{j=1}^{M}$ be the particle set at time step $t$. The particles can be propagated $h \ (h\in \mathbb{Z}^{+})$ using the motion model (\ref{eq: motionmodel}), i.e., 
\begin{equation}
\begin{aligned}
    \bm{X}_{t,h}^{(j)}
    &= \bm{P}^{h}\bm{X}_{t}^{(j)} 
    + \bm{P}^{h-1}\bm{Q}\bm{a}_{t}^{(j)}\\
    &+ \bm{P}^{h-2}\bm{Q}\bm{a}_{t+1}^{(j)}
    + \dots
    + \bm{Q}\bm{a}_{t+h-1}^{(j)}\\
    &= \bm{P}^{h}\bm{X}_{t}^{(j)} 
    + \sum_{k=0}^{h-1}\bm{P}^{k}\bm{Q}\bm{a}_{t+h-1-k}^{(j)}.
\end{aligned}
\end{equation}
Note that this propagation occurs at time $t$. According to the procedures in particle filtering, we have
\begin{equation}
    \bm{X}_{t,1}^{(j)} = \bm{X}_{t+1}^{(j)},\ 
    \bm{X}_{t,h}^{(j)} \ne \bm{X}_{t+h}^{(j)},\ 
    \forall\ h>1.
\end{equation}

The PF predictor and APF predictor then provide $h$-step-ahead point predictions for $\bm{x}_{t+h}$, the target position at time $t+h$:
\begin{equation}
    f_{PF}\left(\{\bm{x}_{t,h}^{(j)}\}_{j=1}^{M}\right) = \hat{\bm{x}}_{t,h},
\end{equation}
\begin{equation}
    f_{APF}\left(\{\bm{x}_{t,h}^{(j)}\}_{j=1}^{M}\right) = \hat{\bm{x}}_{t,h}.
\end{equation}
Similarly, we define a time-varying calibration set at time step $t$ for the $h$-step-ahead conformal inference:
\begin{equation}\label{eq: calibrationseth}
    \mathcal{C}_{t,h}(b) :=
    \left\{
   \mathcal{P}_{t,h}(b),\ (\bm{x}_{t-b+1:t}^{(j)}, w_{t-b+1:t}^{(j)})
   \right\}_{j=1}^{M},
\end{equation}
where
\begin{equation}
    \mathcal{P}_{t,h}(b) := 
    \left\{
    \hat{\bm{x}}_{k,h}: \ t-h-b+1 \le k \le t-h
    \right\}.
\end{equation}
The definition (\ref{eq: calibrationseth}) is equivalent to (\ref{eq: calibrationset1}) when $h=1$. In the same way, letting $\delta_{k,h}^{(j)} = C(\hat{\bm{x}}_{k,h}, \bm{x}_{k+h}^{(j)})$ for $t-h-b+1\le k \le t-h$ and $1\le j \le M$, we define
\begin{equation}
    \hat{\mathcal{Q}}_{t,h}(q) 
    := \inf\left\{
    c: \sum_{k,j}
    \mathbb{I}\left(\delta_{k,h}^{(j)} \le c \right)
    \cdot \frac{w_{k+h}^{(j)}}{b}
    \ge q
    \right\},
\end{equation}
The $h$-step-ahead predicted set is given by
\begin{equation}\label{eq: multisteppredictedset}
    \hat{\mathcal{X}}_{t,h}(\alpha_{t,h}) :=
    \left\{\bm{x}: 
    C(\hat{\bm{x}}_{t,h}, \bm{x}) \le \hat{\mathcal{Q}}_{t,h}(1-\alpha_{t,h})
    \right\},
\end{equation}
and the online update formula is
\begin{equation}
    \alpha_{t+1,h} = \alpha_{t,h} + \gamma\left(\alpha - \Lambda_{t,h}(\alpha_{t,h})\right),
\end{equation}
where $\Lambda_{t,h}(\alpha_{t,h})$ is the realized aggregated miscoverage rate of the predicted set $\hat{\mathcal{X}}_{t,h}(\alpha_{t,h})$, defined as
\begin{equation}
    \Lambda_{t,h}(\alpha_{t,h}) 
    = \sum_{j=1}^{M} w_{t+h}^{(j)} \cdot 
    \mathbb{I}\left(
    \delta_{t,h}^{(j)}>\hat{\mathcal{Q}}_{t,h}(1-\alpha_{t,h})
    \right).
\end{equation}
The similar properties as (\ref{eq: property1}) and (\ref{eq: property2}) are evident.

\section{Simulation Study}
\label{sec: simulationstudy}
To evaluate the effectiveness of our adaptive conformal inference framework, we design a real-time target localization simulation study based on the model structure outlined in the previous sections. This includes simulations for both one-step and multi-step inference. In Section \ref{subsec: simulationsetup}, we first introduce our general simulation setup, followed by an explanation of the key performance metrics utilized in our simulations and the comparison baselines. Finally, in Section \ref{subsec: results}, we present the results from two simulation scenarios—one-step-ahead and multi-step-ahead inferences—evaluating and analyzing performance using the designated metrics.

\subsection{Simulation Setup}
\label{subsec: simulationsetup}
\begin{figure}[ht]
\begin{center}
\centerline{\includegraphics[width=\columnwidth]{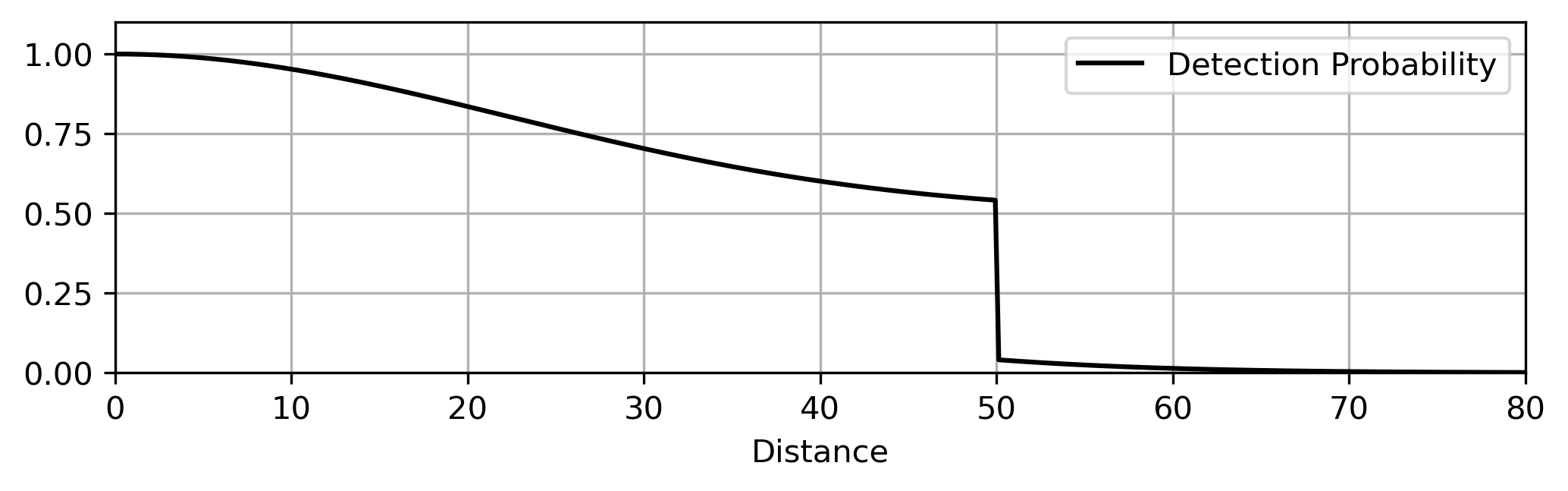}}
\caption{The detection probability of sensors.}
\label{fig: detection probability}
\end{center}
\end{figure}

Recall the target motion model (\ref{eq: motionmodel}). The general settings for our simulation are as follows: the initial hidden state of the target is set to $\bm{X}_{1}=(0,0,1,1)'$, which indicates an initial position of $\bm{x}_1=(0,0)'$ and an initial velocity of $\bm{v}_1=(1,1)'$. The covariance matrix of the target's acceleration is defined as $\Sigma = \text{diag}(0.1,0.1)$. Trajectories are generated over a total of $T=1000$ time steps, with the first $T_0 = 200$ time steps designated for training the particle filter. Inferences are conducted over the remaining $T_1 = 800$ time steps. For the sensor network, we set the sensor density $d_s = 0.001$, decay parameter $\beta=0.001$, strong signal range $r_0=50$, basic detection probability $p_0=1$, and weight parameter $w=0.5$. This configuration yields a detection probability pattern, as illustrated in Figure \ref{fig: detection probability}. After generating the trajectories, sensors are randomly deployed within rectangular maps that encompass the trajectories, maintaining the specified density $d_s = 0.001$. Additionally, the size of the particle set is set to $M=1000$, the look-back parameter in the calibration sets $\mathcal{C}_{t}(b)$ and $\mathcal{C}_{t,h}(b)$ is $b=10$, the step size for adaptive conformal inference is $\gamma=0.01$, and the aggregated miscoverage level is set to $\alpha=0.1$ (aggregated coverage level $1-\alpha=0.9$). 

\subsubsection{Performance Metrics}
We aim for the size of the predicted set at each time step to be as small as possible. Since each predicted set represents a circle in our conformal inference framework, its size is measured by the area. Simultaneously, we expect the realized aggregated coverage (or miscoverage) rate to maintain the level of $1-\alpha$ $(\alpha)$. Therefore, the realized aggregated coverage rate at each time step is crucial. Additionally, although the actual position of the target is unobservable in the online inference framework, we are interested in the actual coverage rate, which measures the ability of the predicted set to encompass the actual position of the target. This allows us to assess how well the aggregated distribution in (\ref{eq: aggregateddistribution}) approximates the actual posterior distribution or even the true distribution of the target location. To summarize these objectives, we define the following performance metrics:
\begin{itemize}
    \item \textbf{Realized Aggregated Coverage (Miscoverage) Rate:} This is the sum of the weights of particles inside (outside) the predicted set at each time step. Its mean should be close to the set level $1-\alpha$ ($\alpha$).
    \item \textbf{Actual Coverage (Miscoverage) Rate:} This metric measures the frequency of the predicted set containing the actual position. The mean of actual coverage rate is defined as $\sum_{t}c_{t}/(\#\text{ of }c_{t})$, where $c_{t}\in \{0,1\}$ indicates whether the predicted set at time $t$ successfully contains the actual target position ($c_{t}=1$) or not ($c_{t}=0$).
    \item \textbf{Area of Predicted Set:} This metric quantifies the size of the predicted set, which is expected to be minimized while maintaining the realized aggregated coverage (or miscoverage) rate at the level $1-\alpha$ ($\alpha$).
\end{itemize}

\subsubsection{Comparison Baselines}
The adaptive conformal inference method updates the parameter $\alpha_{t}$ to adapt to the time-varying distribution of data, continuously calibrating our estimate of the optimal parameter $\alpha_{t}'$. In order to evaluate the advantages of this adaptive online update strategy, we compare the adaptive conformal inference method with a non-adaptive approach. The latter maintains a constant parameter $\alpha_{t}\equiv \alpha$, without updating $\alpha_{t}$. These non-adaptive models serve as the baseline for comparison.

\subsection{Results}
\label{subsec: results}
In this section, we present the results of our real-time target localization simulation study, aimed at evaluating the effectiveness of the proposed framework. The results include both one-step-ahead and multi-step-ahead online adaptive conformal inferences, which are compared to the non-adaptive baselines using three designed metrics: realized aggregated coverage rate, actual coverage rate, and area of predicted set.

\subsubsection{One-Step Inference Results}
\begin{figure}[htp]
\begin{center}
\centerline{\includegraphics[width=\columnwidth]{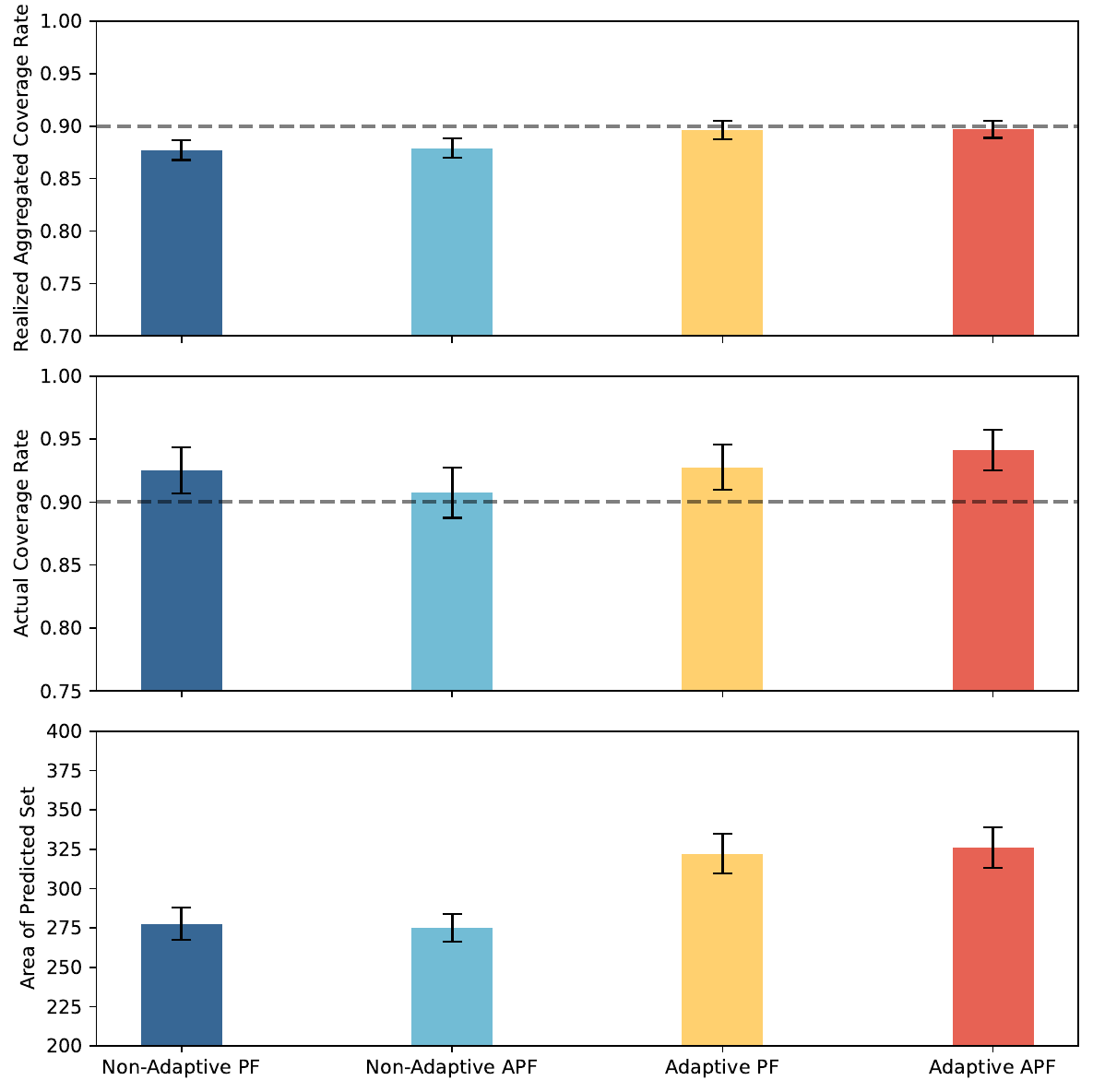}}
\caption{Performance of one-step-ahead conformal inference for target localization using PF and APF predictors, comparing non-adaptive and adaptive update strategies. The dashed line represents the set aggregated coverage level $1-\alpha=0.9$. Error bars indicate the corresponding 95\% confidence intervals.}
\label{fig: onestep_3metrics}
\end{center}
\end{figure}

The one-step-ahead conformal inference is conducted over a total of $T_1=800$ time steps, with performance results displayed in Figure \ref{fig: onestep_3metrics} and Table \ref{tab: onestep_3metrics}. The adaptive inferences using the PF and APF predictors achieve realized aggregated coverage rates of 0.8966 and 0.8971, respectively, both of which have 95\% confidence intervals that encompass the set aggregated coverage level of 0.9. In contrast, the non-adaptive inferences yield slightly lower performance, with corresponding realized aggregated coverage rates of 0.877 and 0.879. It is observed that the area of the predicted set in the adaptive inferences is typically larger than that in the non-adaptive inferences. This suggests that the adaptive update strategy effectively calibrates the estimate of the optimal parameter at each time step, resulting in precise control of the mean aggregated coverage rate.

\begin{table}[htp]
	\caption{Realized aggregated coverage rate (RACR), actual coverage rate (ACR), and area of predicted set (APS) for the one-step-ahead conformal inference results, averaged over all prediction time steps.}
	\label{tab: onestep_3metrics}
	\begin{center}
	\begin{tabular}{lllll} %
		\toprule
		Metrics & (N-A) PF & (N-A) APF & (A) PF & (A) APF \\
		\midrule
		RACR &  0.877 & 0.879 & 0.8966 & 0.8971 \\
  	ACR &  0.925 & 0.9075 & 0.9275 & 0.9412 \\
		APS & 277.4639 & 275.1513 & 321.9845 & 326.0098 \\\bottomrule
	\end{tabular}
	\end{center}
	\footnotesize
	\emph{Note: (N-A) and (A) indicate non-adaptive and adaptive update strategy-based models, respectively.}
\end{table}

Interestingly, all four types of conformal inferences achieve the set level of 0.9 and even exceed it in terms of the actual coverage rate. This discrepancy may arise from differences between the aggregated posterior distribution and the actual posterior distribution, as well as the true distribution. To mitigate the uncertainty associated with a large number of particles in the space, our conformal inference method may exhibit a degree of conservatism concerning the actual target position. However, since the actual target position is unobservable under the current assumptions, this indicates that our conformal inference framework is effective for online target localization, focusing primarily on the aggregated posterior distribution.

\subsubsection{Multi-Step Inference Results}
\begin{figure}[htp]
\begin{center}
\centerline{\includegraphics[width=\columnwidth]{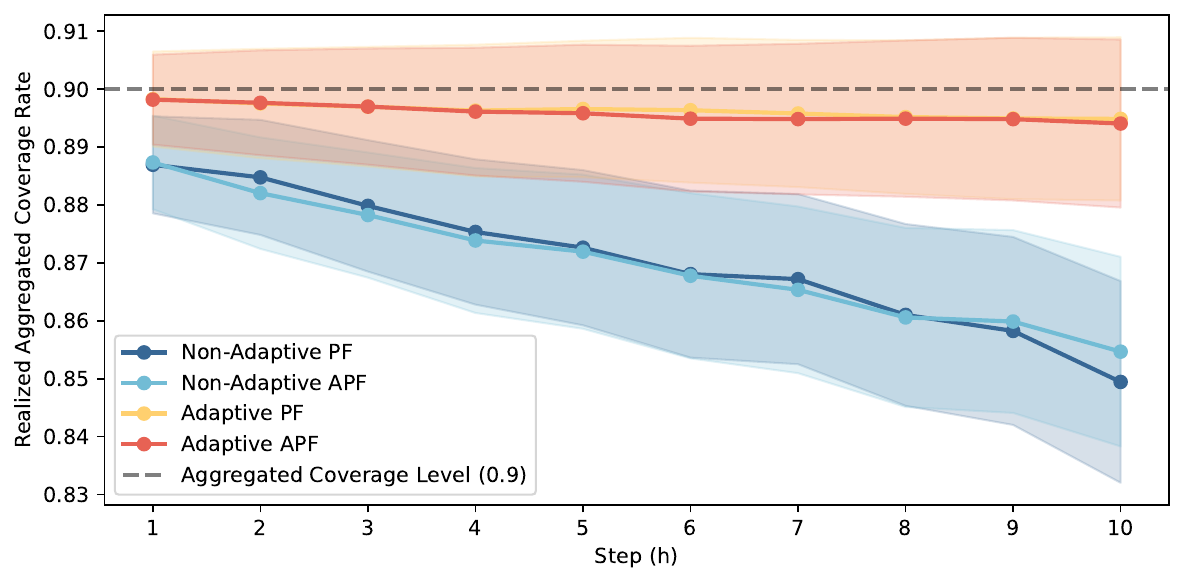}}
\caption{Aggregated coverage rate of multi-step-ahead conformal inference for target localization through PF and APF predictors, comparing non-adaptive and adaptive update strategies. The colored regions represent the corresponding 95\% confidence intervals.}
\label{fig: multistep_aggregated_coverage_rate}
\end{center}
\end{figure}

The total number of time steps for conducting the multi-step-ahead conformal inference is also $T_1=800$. Specifically, we perform a 10-step-ahead conformal inference and evaluate the outcomes at the last 800 time steps. Figure \ref{fig: multistep_aggregated_coverage_rate} illustrates the realized aggregated coverage rates for the $h$-th ($1\le h \le 10$) step inference results across the four types of methods.
As $h$ increases, we observe that the variances of the realized aggregated coverage rates also increase. A larger step $h$ introduces greater uncertainty, resulting in less accurate point predictions by the PF and APF predictors. Additionally, the mean realized aggregated coverage rate decreases with increasing $h$. However, the adaptive conformal inferences maintain the realized aggregated coverage rate close to the set level of 0.9, while the non-adaptive inferences experience a significant decline in their realized aggregated coverage rates.

\begin{figure}[htp]
\begin{center}
\centerline{\includegraphics[width=\columnwidth]{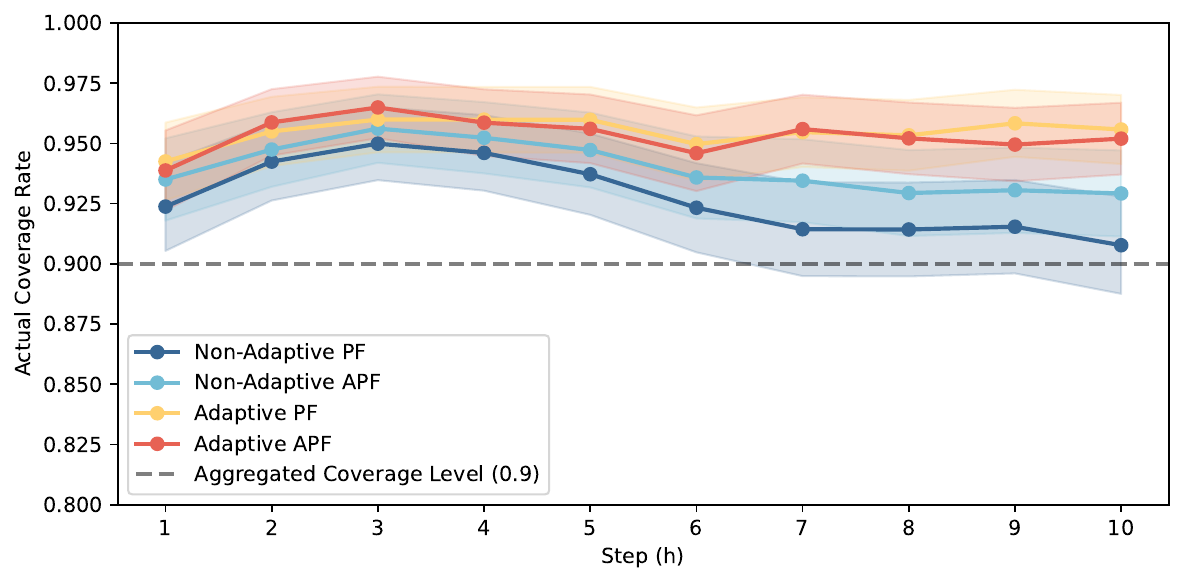}}
\caption{Actual coverage rate of multi-step-ahead conformal inference for target localization through PF and APF predictors, comparing non-adaptive and adaptive update strategies. The colored regions represent the corresponding 95\% confidence intervals.}
\label{fig: multistep_actual_coverage_rate}
\end{center}
\end{figure}

Similar to the one-step-ahead conformal inference results, the actual coverage rates exceed the set realized aggregated coverage rate across all four methods, as shown in Figure \ref{fig: multistep_actual_coverage_rate}. Notably, as the time span ahead $h$ increases, the actual coverage rates of the adaptive conformal inferences stabilize around 0.95, demonstrating greater stability compared to the non-adaptive conformal inferences. This indicates that the adaptive update strategy effectively maintains a stable realized aggregated coverage rate, which in turn results in a relatively stable actual coverage rate.

\begin{figure}[htp]
    \centering
    \begin{minipage}[b]{\linewidth}
        \centering
        \includegraphics[width=\linewidth]{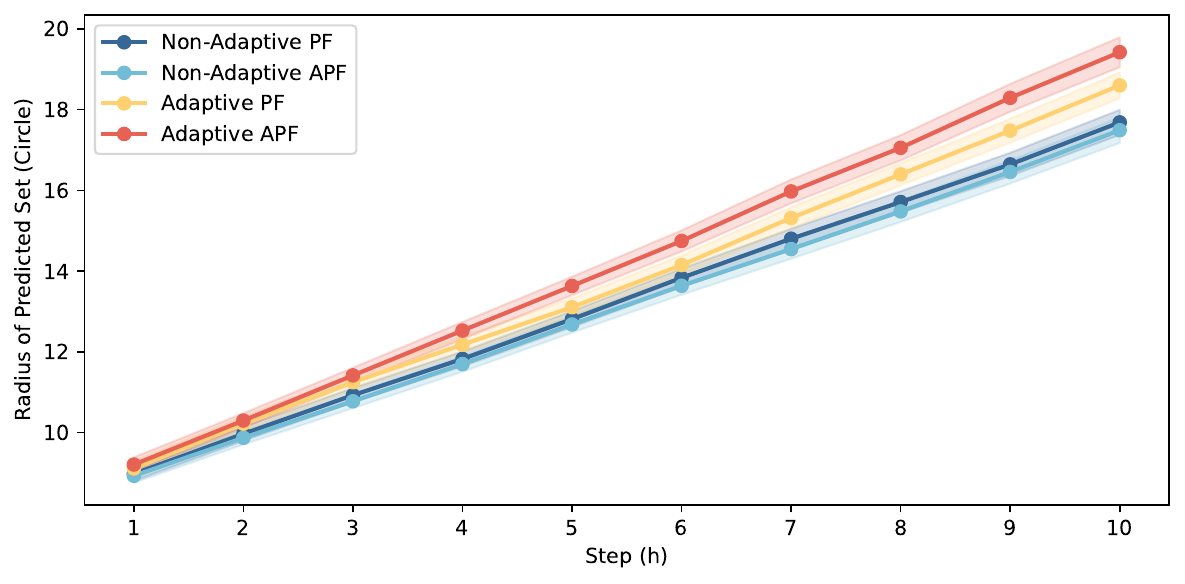}
    \end{minipage}
    \hfill
    \begin{minipage}[b]{\linewidth}
        \centering
        \includegraphics[width=\linewidth]{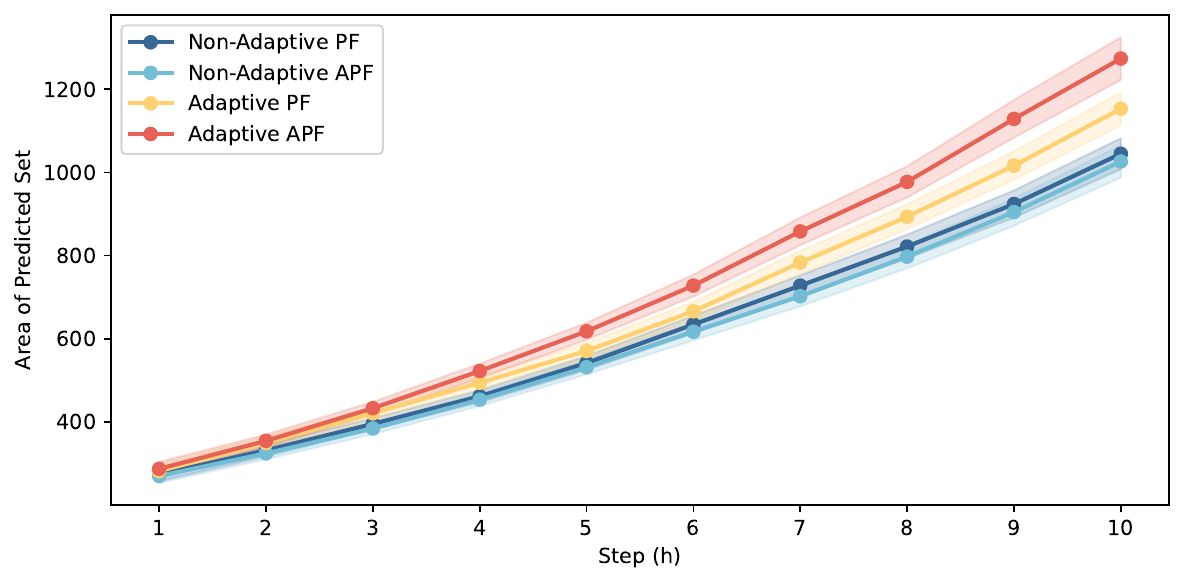}
    \end{minipage}
    \caption{Radius (upper panel) and area (lower panel) of the predicted set from multi-step-ahead conformal inference for target localization using PF and APF predictors, with non-adaptive and adaptive update strategies. The colored regions represent the corresponding 95\% confidence intervals.}
    \label{fig: multistep_predicted_area}
\end{figure}

The PF and APF predictors provide point predictions for the target position following the motion model in (\ref{eq: motionmodel}). However, the particles may not fully capture the characteristics of the target motion, leading to increasing deviations as the time span ahead $h$ grows. To compensate for this deviation, the conformal inferences exhibit a linearly increasing radius of the predicted set with $h$ (see the upper panel in Figure \ref{fig: multistep_predicted_area}), resulting in a quadratic trend for the corresponding area (see the lower panel in Figure \ref{fig: multistep_predicted_area}). We observe that the adaptive inferences consistently yield slightly larger set sizes compared to the non-adaptive inferences, correlating with higher realized aggregated coverage rates that remain close to the set level.

\begin{figure}[htp]
\begin{center}
\centerline{\includegraphics[width=\columnwidth]{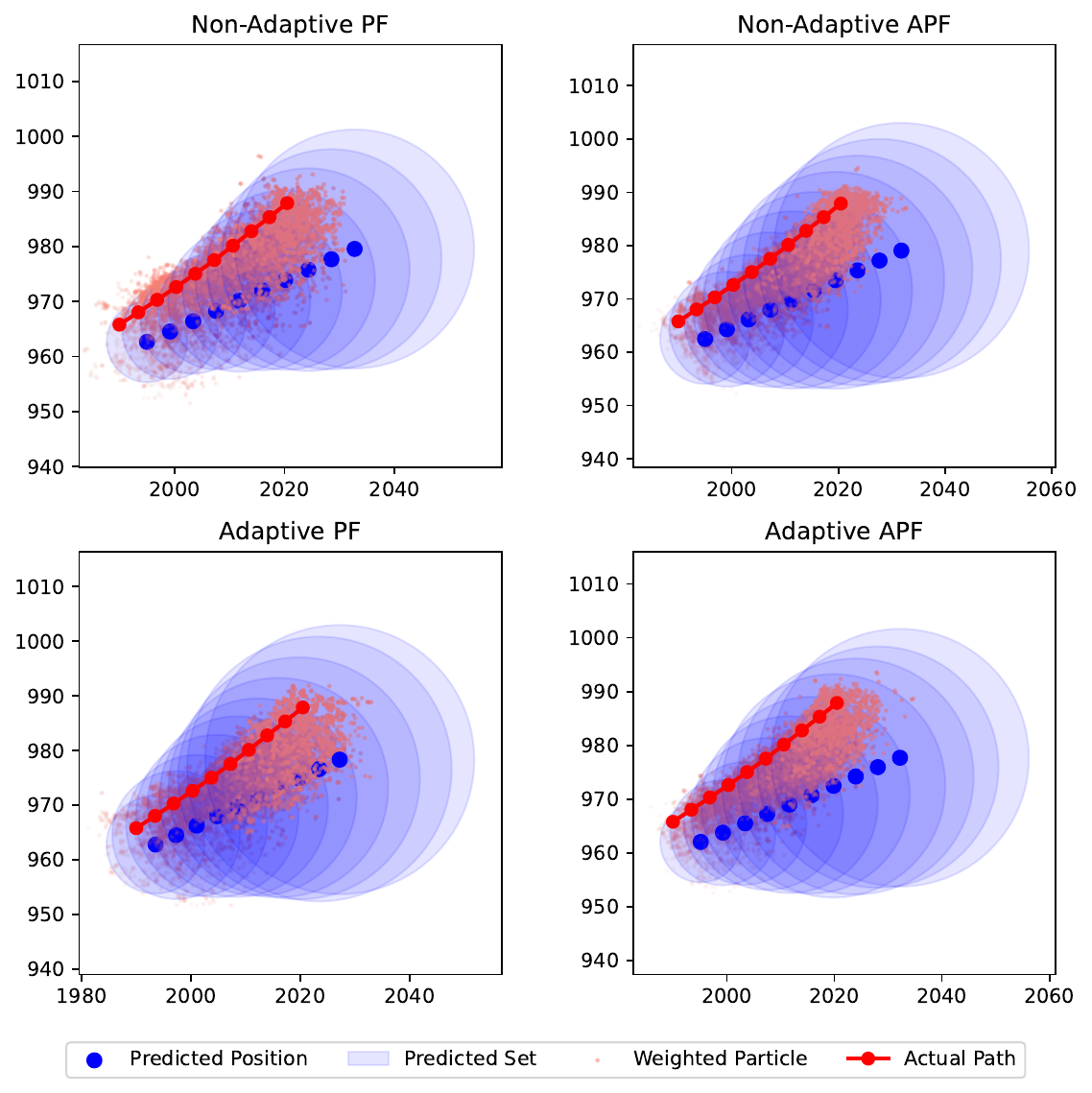}}
\caption{10-step-ahead conformal inference for target localization using PF and APF predictors, comparing non-adaptive (upper panel) and adaptive (lower panel) update strategies.}
\label{fig: multistep_lateststeps}
\end{center}
\end{figure}

We present a realization of 10-step-ahead conformal inferences in Figure \ref{fig: multistep_lateststeps}. As $h$ increases from 1 to 10, the size of the predicted set becomes larger, and the predicted position becomes more inaccurate. Numerous weighted particles are distributed around the actual path of the target, demonstrating that particle filtering is an effective tool for approximating the posterior distribution of the target position. The predicted sets generated by the adaptive conformal inferences encompass more weighted particles than those produced by the non-adaptive method, particularly with the PF predictor, as shown in the upper left and lower left panels of Figure \ref{fig: multistep_lateststeps}. Although the weighted particle set appears more concentrated when using auxiliary particle filtering, we do not assert that the PF predictor is superior to the APF predictor within our conformal inference framework; both predictors yield similarly satisfactory results. Instead, our goal is to highlight the effectiveness of the proposed conformal inference framework, which focuses on the aggregated posterior distribution, specifically the weighted particles provided by particle filtering.

\section{Discussion}
\label{sec: discussion}
Throughout the real-time target localization simulation study, we have demonstrated the effectiveness of the proposed framework applied to one-step and multi-step online conformal inference. The adaptive update strategy provides asymptotically precise control of the realized aggregated coverage rate by continuously calibrating the errors in optimal parameter estimation. Focusing on the aggregated posterior distribution effectively characterizes the actual distribution of the target position, resulting in satisfactory actual coverage rates. 

By setting the aggregated coverage level at $1-\alpha=0.9$, the adaptive conformal inferences with PF and APF predictors achieve realized aggregated coverage rates that are very close to this set level, and they outperform the actual coverage rates in the one-step-ahead inference simulations (see Figure \ref{fig: onestep_3metrics} and Table \ref{tab: onestep_3metrics}). In multi-step inferences, the realized aggregated coverage rates remain robust against the uncertainty introduced by the time span ahead $h$, exhibiting significant stability in both realized aggregated and actual coverage rates across varying $h$ (see Figure \ref{fig: multistep_aggregated_coverage_rate} and Figure \ref{fig: multistep_actual_coverage_rate}). Furthermore, to compensate for the uncertainty and deviation in point predictions, the size of the predicted set increases with $h$, demonstrating a linear trend in the radius and a quadratic trend in the area (see Figure \ref{fig: multistep_predicted_area} and Figure \ref{fig: multistep_lateststeps}).

However, the proposed framework does have limitations. One concern is that the chosen update step size $\gamma$ might significantly affect the adaptive response to the time-varying distribution. The theoretical results in \cite{Gibbs2021} suggest that $\gamma$ should be proportional to the variation size of the optimal parameter $\alpha_{t}'$ over time, which is typically unknown. The work by \cite{Gibbs2024} introduced the dynamically-tuned adaptive conformal inference, which is a modified version of an algorithm proposed by \cite{Gradu2023}. This approach runs multiple versions of adaptive conformal inference in parallel with a set of candidate $\gamma$ values and the corresponding $\alpha_{t}$ values, selecting the  $\alpha_{t}$ at each time by evaluating the historical performance of these versions. Our framework could be enhanced by incorporating this consideration. Additionally, the point predictions generated by the PF and APF predictors may not be optimal, as the deviation typically grows with the time span ahead $h$, leading to a quadratic increase in the area of the predicted set. This conservatism in inference is expected to be reduced.

Overall, we emphasize the potential and effectiveness of utilizing the aggregated posterior distribution derived from the weighted particles in particle filtering as an approximation of the actual posterior distribution in online adaptive conformal inference. Future work could explore improvements in the online adaptive update strategy, the selection of predictors, and methods for constructing predicted sets in conformal inference, as well as specific approaches for different environments characterized by rapidly shifting data distributions.

\section{Conclusion}
\label{sec: conclusion}
This paper proposes an online adaptive conformal inference framework for hidden states under the assumption of hidden Markov models (HMMs). To address the challenge of unobservable hidden states, the framework innovatively utilizes numerous weighted particles generated by particle filtering as an approximation of the actual posterior distribution of the hidden state within the context of conformal inference. By defining the aggregated coverage rate as the sum of the weights of particles contained in the prediction sets, the proposed method aims to create prediction sets that effectively cover these particles to achieve a specified aggregated coverage level. A real-time target localization simulation study demonstrates the effectiveness of the proposed framework, showcasing its stable control of the realized aggregated coverage rate in both one-step and multi-step inferences.

\section{Acknowledgment}
\label{sec: acknowledgment}
The work described in this paper was supported by grants from 
City University of Hong Kong (Project No. 9610639 and No. 6000864 ).

\end{document}